# PREDICTION OF CRYPTOCURRENCY PRICES USING LSTM, SVM AND POLYNOMIAL REGRESSION

**Novan Fauzi Al Giffary, Feri Sulianta**
Widyatama University, Indonesia
Email: novan.fauzi@widyatama.ac.id, feri.sulianta@widyatama.ac.id

## ABSTRACT
The rapid development of information technology, especially the Internet, has facilitated users with a quick and easy way to seek information. With these convenience offered by internet services, many individuals who initially invested in gold and precious metals are now shifting into digital investments in form of cryptocurrencies. However, investments in crypto coins are filled with uncertainties and fluctuation in daily basis. This risk posed as significant challenges for coin investors that could result in substantial investment losses. The uncertainty of the value of these crypto coins is a critical issue in the field of coin investment. Forecasting, is one of the methods used to predict the future value of these crypto coins. By utilizing the models of Long Short Term Memory, Support Vector Machine, and Polynomial Regression algorithm for forecasting, a performance comparison is conducted to determine which algorithm model is most suitable for predicting crypto currency prices. The mean square error is employed as a benchmark for the comparison. By applying those three constructed algorithm models, the Support Vector Machine uses a linear kernel to produce the smallest mean square error compared to the Long Short Term Memory and Polynomial Regression algorithm models, with a mean square error value of 0.02.

## INTRODUCTION

The rapid advancement of information technology, particularly the internet, has exponentially accelerated on a daily basis, providing users with expedited access to information (Gill et al., 2024; Mathkor et al., 2024). Leveraging the convenience offered by internet services, many individuals who previously invested their funds in physical assets such as gold and precious metals are transitioning towards digital investments, exemplified by cryptocurrencies.

However, investments in crypto currency are subject to daily fluctuations, posing substantial challenges and potential financial losses for coin investors. The uncertainty surrounding the fluctuations in the prices of cryptocurrencies represents a critical concern within the coin investment landscape, necessitating the development of predictive models for crypto currency prices. In order to address this need, various methodologies such as Long-Short-Term Memory (LSTM), Support Vector Machine, and Polynomial Regression are employed for prediction purposes (Jana et al., 2023; Kalu et al., 2023; Yao et al., 2017). Long-Short-Term Memory (LSTM), commonly known as LSTM, is a type of Artificial Neural Network (ANN) utilized in deep learning and artificial intelligence domains (Sabry, 2023). Recognized for its ability to retain information over extended periods, LSTM is particularly suitable for time series data. Support Vector Machine (SVM) is widely acknowledged for its application in automated classification, lending itself to predictive. Polynomial Regression, on the other hand, represents an extension of Linear Regression, offering predictive capabilities tailored to address non-linear data patterns (Mordensky et al., 2023; Murugan et al., 2023; Wangwongchai et al., 2023).

a prior study by Wei Cao, historical Tesla stock data was used to compare the Long Short Term Memory (LSTM) method with linear regression (Cao, 2021). The research utilized batch variables and Tesla stock historical data from June 2010 until February 2020 to assess the suitability of each method in predicting Tesla stock prices (Radojičić & Kredatus, 2020; Zhao et al., 2023). Mean Square Error (MSE) served as the benchmark, revealing that the LSTM method exhibited lower MSE compared to linear regression. On the other hand, Dong Liu, Ang Chen, and Juanjuan Wu (Cao, 2021) employed Recurrent Neural Network (RNN) and LSTM to predict stock prices using GREE data from the Yahoo Finance database. The study covered 728 days of stock data from Yahoo Finance, evaluating the two methods with MSE and MAE. Results indicated that the LSTM model produced a smaller MSE compared to the RNN model (Aseeri, 2023; Gülmez, 2023).

Felix Zhan (Zhan, 2021) used Polynomial Regression to predict the number of COVID-19 cases in the United States, using datasheets from the COVID Tracking Project and the United States Census Bureau. The DyCPR algorithm was employed to predict H+1 active COVID-19 cases. Results after training and testing showed that DyCPR was more accurate in predicting stable time series. Zixuan Liu, Ziyuan Dang, and Jie Yu (Liu et al., 2020) employed the Support Vector Machine (SVM) algorithm to predict stock prices using historical data from the Yahoo Finance database. Evaluating model accuracy with various algorithms such as RBFSVM, PCA-SVM, GA-SVM, and DFS-BPSO, training results revealed that the RBF-SVM algorithm outperformed others in accuracy. Based on these studies, this research aims to analyze crypto currency price predictions using the Long Short Term Memory, Support Vector Machine, and Polynomial Regression methods. Additionally, the study aims to compare the performance of these three methods using Mean Square Error as a benchmark variable.

## RESEARCH METHODS

This research was conducted in several stages, including data collection, data pre-processing, data allocation, algorithm model design, data training, data testing, and result evaluation by comparing three algorithms using the mean square error variable. The programming language utilized in this research was Python. The procedural flow of these research stages can be observed in Figure 1.

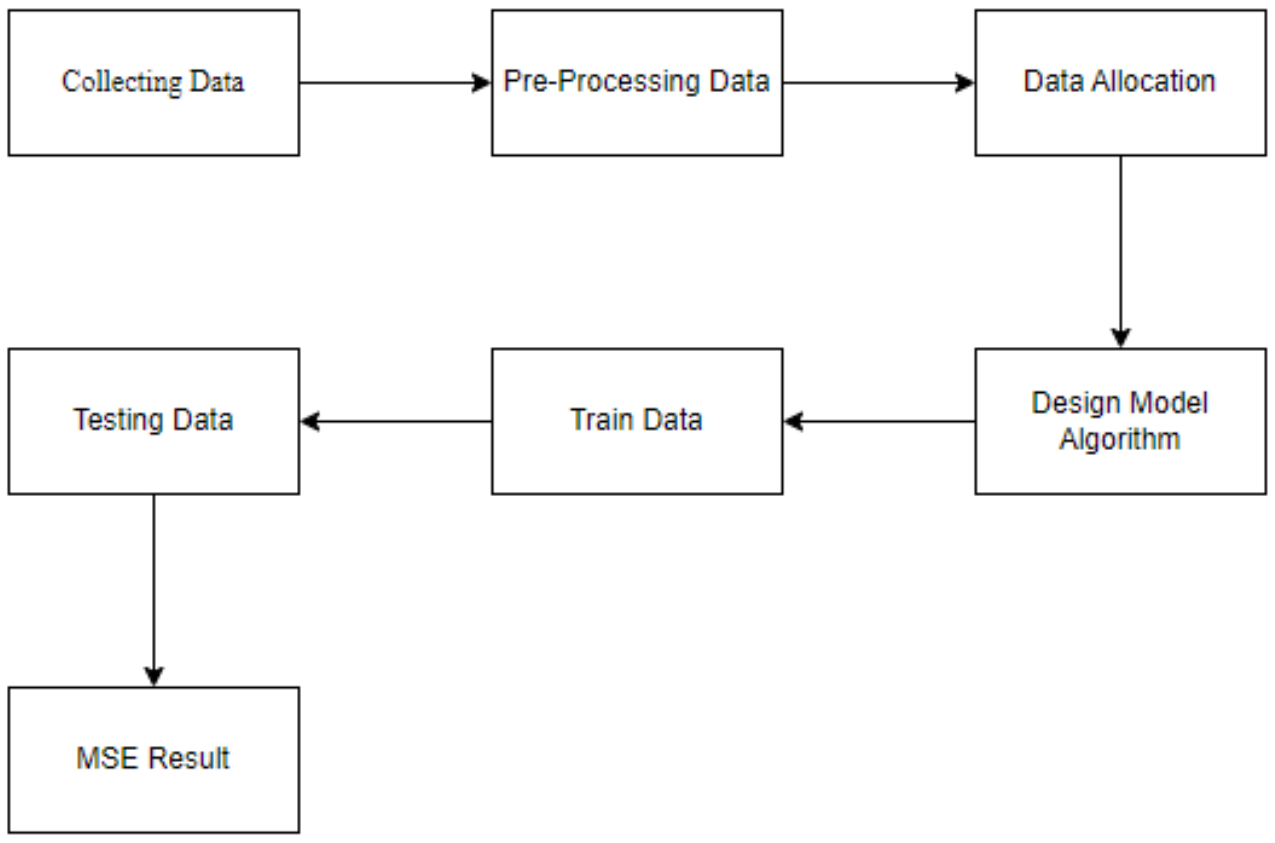

**Figure. 1 The stages of the research**

In this research, the proposed Algorithm method involves the utilization of three algorithm models: Long Short Term Memory (LSTM), Support Vector Machine (SVM), and Polynomial Regression. These Algorithm models serve as experimental components, with the mean square error variable employed as the determinant. Long Short Term Memory is acknowledged for its proficient predictive modelling capabilities and is derived from Recurrent Neural Network, a method designed for processing sequence data (Rizkilloh & Widiyanesti, 2022). Some researchers have employed Support Vector Machine for predicting student graduation times (Haryatmi & Hervianti, 2021). Polynomial Regression, on the other hand, is effective in handling non-linear data patterns (Di Lallo et al., 2023; Helin et al., 2023; Reza et al., 2023).

In detail, the experimental process divided into several stages. It commences with data collection, wherein the research data comprises Bitcoin price data from 2020 to 2024 obtained from the Yahoo Finance website. Subsequently, the dataset undergoes normalization to eliminate null values, minimizing errors during algorithm model testing. After normalization, the data allocated into two categories: training data and testing data, with an 80:20 ratio for training and testing, respectively. This ratio signifies that 80% of the data will be used for training, while 20% will be utilized for testing the data.

**Table 1. Data Sample**

| Date | Open | High | Low | Close | Adj Close | Volume |
|---|---|---|---|---|---|---|
| 10/01/2020 | 7878.3076 | 8166.5541 | 7726.7749 | 8166.5541 | 8166.5541 | 28714583844 |
| 11/01/2020 | 8162.1909 | 8218.3593 | 8029.6420 | 8037.5375 | 8037.5375 | 25521165085 |
| 12/01/2020 | 8033.2617 | 8200.0634 | 8009.0590 | 8192.4941 | 8192.4941 | 22903438381 |
| 13/01/2020 | 8189.7719 | 8197.7880 | 8079.7006 | 8144.1943 | 8144.1943 | 22482910688 |
| 14/01/2020 | 8140.9331 | 8879.5117 | 8140.9331 | 8827.7646 | 8827.7646 | 44841784107 |
| 15/01/2020 | 8825.3437 | 8890.1171 | 8657.1875 | 8807.0107 | 8807.0107 | 40102834650 |

After the data is allocated, model testing is conducted, divided into three Algorithm models. The Long Short Term Memory algorithm model, in particular, incorporates multiple hidden layers. These layers function during the flow of information, where each pertinent piece of information is retained, and irrelevant information is discarded in each cell, this algorithm has the architecture shown in Figure 2 (Hasiholan et al., 2022).

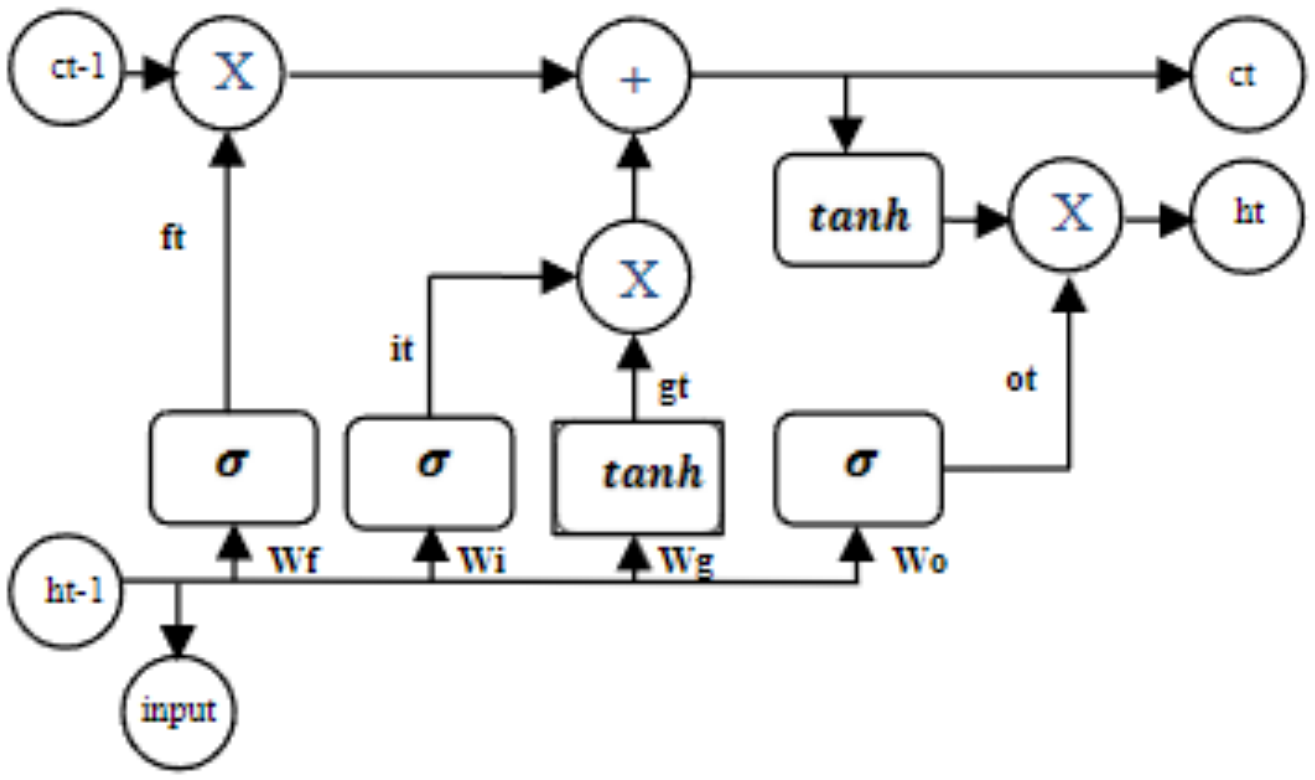

**Figure. 2 Long Short Term Memory algorithm**

In the Long Short Term Memory algorithm has the equation:

Input gate:

$i_{(t)}=\sigma(x_t \cdot W_{ix}+ h_{(t-1)} \cdot W_{ih}+ b_i)$ (1)

Forget gate:

$f_{(t)}=\sigma(x_t \cdot W_{fx}+ h_{(t-1)} \cdot W_{fh}+ b_f)$ (2)

Candidate Cell State:

$$\tilde{C}_{t} = \tanh(x_t \cdot W_{cx} + h_{t-1} \cdot W_{ch} + b_c) \quad (3)$$

Output Gate:

$$o_{t} = \sigma(x_t \cdot W_{ox} + h_{t-1} \cdot W_{oh} + b_o) \quad (4)$$

Cell State:

$$C_{t} = f_t \circ C_{t-1} + i_t \circ \tilde{C}_t \quad (5)$$

Hidden State:

$$h_{t} = o_t \circ \tanh(C_t) \quad (6)$$

In the Long Short Term Memory algorithm model, the Adam optimizer algorithm is employed for network optimization and accuracy enhancemen (Tan et al., 2018). For the training data, the training process is iterated multiple times with varying Epoch numbers specifically, 10, 30, 50, 80, and 100, to achieve the best accuracy results.

On the other hand, in the Support Vector Machine algorithm model, the linear Support Vector Machine can be extended to address non-linear problems. The linear Support Vector Machine can transform into a non-linear Support Vector Machine utilizing kernel methods (Damasela et al., 2022). SVM was utilize to find the best hyperplane by maximizing the distance between classes. The hyperplane is a function used for class separation, referred to as a line in 2-D classification, a plane in 3-D classification, and a hyperplane in higher-dimensional class space.

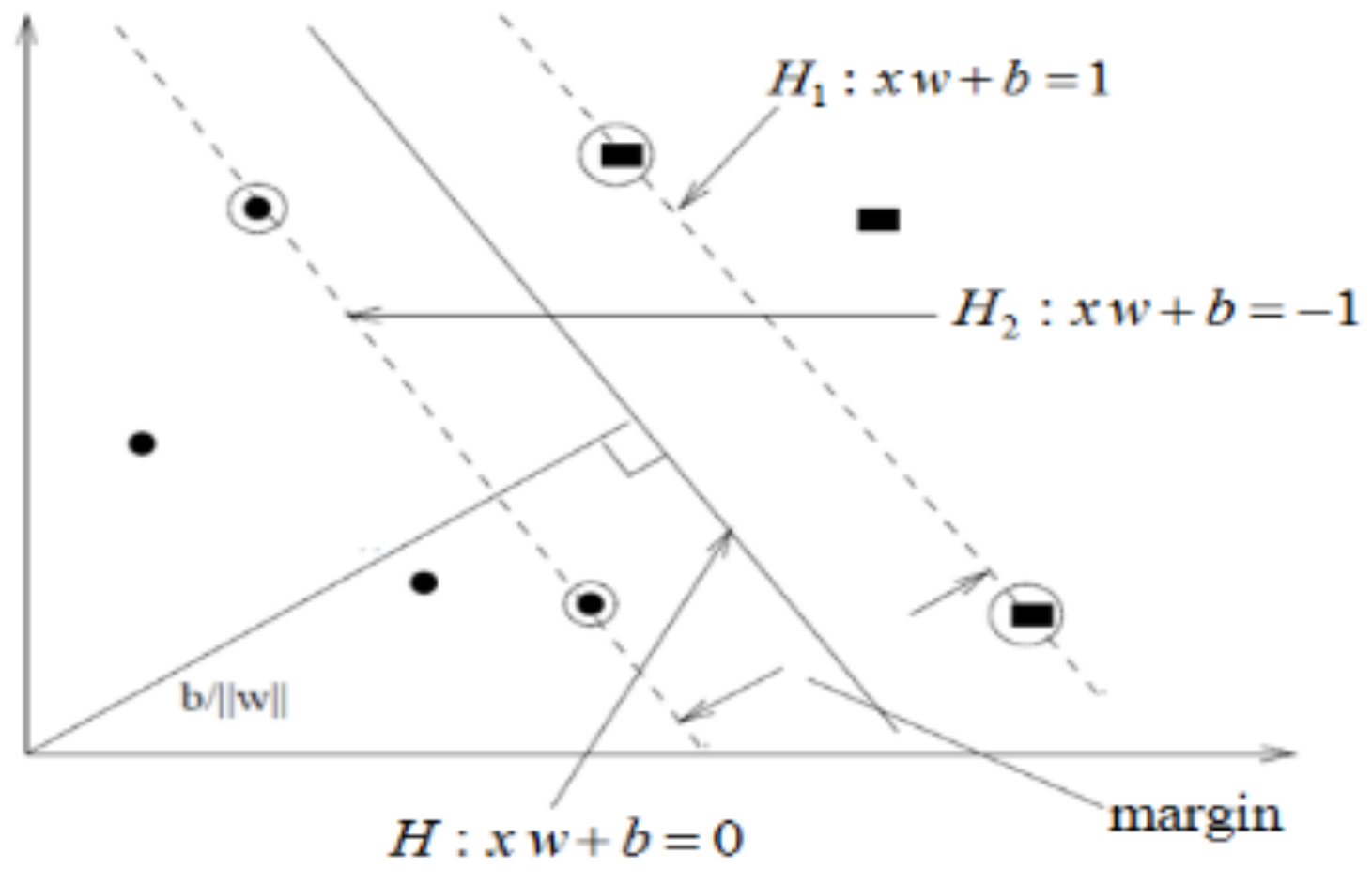

**Figure. 3 Optimal hyperplane that separates two classes**

The Support Vector Machine algorithm model uses Radial Base Function (RBF), Sigmoid, and Linear kernel types with the regularization parameter C set to 1e0, 1e1, 1e2, or 1e3, and the gamma kernel coefficient is set to 0.001, 0.01, 0.1 or 1. The GridSearchCV tuning method also utilized, where the model is constructed by evaluating combinations of parameters. This process aim is to assess the combined results of the C parameter and gamma coefficient used in this model. As for the Polynomial Regression algorithm model, it is an extension of Linear Regression designed to handle non-linear data patterns, making it suitable for predictive purposes. In Polynomial Regression, the degree parameter is employed. Regression degree or parameter degree was employed as an analytic solution for the optimization problems (Antille & Allen, 2022). On this model, the degree parameter sets to 2, 4, 6, 9, and 11. The pre-processed dataset then trained in the constructed Algorithm models, and data testing is performed. The output of the data testing is compared with the output from the training data, generating the mean square error value The general form of Polynomial Regression expressed (Mustafidah & Rohman, 2023) as:

$$y=a^0+a_1 x+a_2 x^2+a_3 x^3+\cdots+a_n x^n \quad (7)$$

For the error equation, the formulation according to (Mustafidah & Rohman, 2023) is:

$$D^2=\sum_{i=1}^{n} \left[ (y_i-a_0-a_1 x-a_1 x^2-\ldots-a_n x^n) \right] \quad (8)$$

The mean square error serves as the benchmark variable in this study, where a lower mean square error indicates a better predictive result.

## RESULTS AND DISCUSSION

During the duration of this research, the analysis was carried out using a prepared dataset from 2020 to 2024 containing information on the price of the Bitcoin currency. The key variable used for predictive modeling is the detailed closing price data, as well as using the Mean Square Error value as a comparison of which algorithm models are more suitable for predicting Bitcoin prices.

**Table 2. Test results from the Long Short Term Memory algorithm model.**

| Epoch | MSE |
|---|---|
| 10 | 398.74313739554805 |
| 30 | 97.91950725856172 |
| 50 | 299.68280456164393 |
| 80 | 132.83835875171226 |
| 100 | 105.23378629880129 |

From the test result of Long Short Term Memory algorithm model, in table 1 it is clear that the epoch count yielding the lowest mean square error is observed at 30 epochs when compared to the others. Meanwhile the highest mean square error value was at 10 epochs, at every increase in the epochs the mean square error value does not always become lower.

**Table 3. Test results from the Support Vector Machine algorithm model**

| Kernel | Gamma | C | MSE |
|---|---|---|---|
| Radial Base Function | 0.001 | 1e0 | 243055088.87 |
| | | 1e1 | 240811202.02 |
| | | 1e2 | 218948815.13 |
| | | 1e3 | 58870606.87 |
| | 0.01 | 1e0 | 241178058.07 |
| | | 1e1 | 222470791.54 |
| | | 1e2 | 78175331.12 |
| | | 1e3 | 1830619.78 |
| | 0.1 | 1e0 | 236345016.21 |
| | | 1e1 | 179439518.42 |
| | | 1e2 | 25529293.18 |
| | | 1e3 | 2185936.11 |
| | 1 | 1e0 | 240347675.94 |
| | | 1e1 | 215899006.69 |
| | | 1e2 | 88797294.67 |
| | | 1e3 | 9483829.60 |
| Sigmoid | 0.001 | 1e0 | 243177530.52 |
| | | 1e1 | 242031314.54 |
| | | 1e2 | 230719546.13 |
| | | 1e3 | 132840755.73 |
| | 0.01 | 1e0 | 242034936.92 |
| | | 1e1 | 230754799.37 |
| | | 1e2 | 133094705.92 |
| | | 1e3 | 351921.13 |
| | 0.1 | 1e0 | 232728594.72 |
| | | 1e1 | 148402089.73 |
| | | 1e2 | 13173250.76 |
| | | 1e3 | 325838418.42 |
| | 1 | 1e0 | 218892924.21 |
| | | 1e1 | 108781933.83 |
| | | 1e2 | 240862304.99 |
| | | 1e3 | 13703086399.99 |
| Linear | 0.001 | 1e0 | 132838133.04 |
| | | 1e1 | 348911.58 |
| | | 1e2 | 156426.94 |
| | | 1e3 | 0.02 |
| | 0.01 | 1e0 | 132838133.04 |
| | | 1e1 | 348911.58 |
| | | 1e2 | 156426.94 |
| | | 1e3 | 0.02 |
| | 0.1 | 1e0 | 132838133.04 |
| | | 1e1 | 348911.58 |
| | | 1e2 | 156426.94 |
| | | 1e3 | 0.02 |
| | 1 | 1e0 | 132838133.04 |
| | | 1e1 | 348911.58 |
| | | 1e2 | 156426.94 |
| | | 1e3 | 0.02 |

Meanwhile, the testing of the support vector machine algorithm model involves a five - fold cross validation for testing in this test. Results of the test in table 3, the kernel that yielded the smallest mean square error was linear kernel with gamma 0.001, 0.01, 0.1 and 1 and parameter c 1e3, while the kernel resulting in the largest mean square error was the sigmoid kernel with gamma 1 and parameter c 1e3.

**Table 4. Test results from the Polynomial Regression Algorithm model**

| Degree | MSE |
|---|---|
| 2 | 161060413.70 |
| 4 | 72475644.19 |
| 6 | 51702001.51 |
| 9 | 172168041.06 |
| 11 | 218284846.10 |

During the testing of the Polynomial Regression Algorithm model, with the parameter "include bias" set to false, it is observed that the degree value producing the lowest mean square error is degree 6. Meanwhile the highest mean square error is degree 11, In each addition of degree the mean square error value produced is always different and does not always become lower.

The comprehensive results of the experiments have been obtained, providing a clearer understanding of each Algorithm model employed. Additionally, the comparison allows us to identify which Algorithm model is more suitable for predicting Bitcoin prices.

**Table 5. comparative analysis of each algorithm model.**

| algorithm model | MSE |
|---|---|
| Long Short Term Memory | 97.91950725856172 |
| Support Vector Machine | 0.02 |
| Polynomial Regression | 51702001.51 |

From Table 5, it is clear that each algorithm produces distinct mean square error values, even when utilizing the same dataset. For instance, the Long Short Term Memory algorithm model yields the lowest mean square error with a value of 97.91950725856172, while the Support Vector Machine algorithm model records the lowest mean square error at 0.02. Similarly, Polynomial Regression yields the lowest mean square error with a value of 51702001.51. However, it is crucial to note that crypto currency prices are not solely confined to historical data, and various factors can influence the pricing dynamics of cryptocurrencies.

## CONCLUSION

Among all three Algorithm models tested, the Support Vector Machine algorithm model shows the lowest mean square error value at 0.02. Meanwhile, the Polynomial Regression algorithm model produced the highest mean square error amount at 51702001.51. Therefore, it is concluded that the Support Vector Machine algorithm model yields the lowest mean square error when compared to the Long Short Term Memory and Polynomial Regression algorithm models for crypto currency price prediction in this study. For future research endeavours, it is advised to include additional variables that may impact prediction outcomes, such as sentiment analysis from individuals. That could be considered as the cause of fluctuations in crypto currency prices.